\documentclass{article}

\PassOptionsToPackage{numbers, compress}{natbib}

     \usepackage[final]{tackling_climate_workshop_style}

\usepackage[utf8]{inputenc} 
\usepackage[T1]{fontenc}    
\usepackage{hyperref}       
\usepackage{url}            
\usepackage{booktabs}       
\usepackage{amsfonts}       
\usepackage{nicefrac}       
\usepackage{microtype}      
\usepackage{amsmath}
\usepackage[capitalize,noabbrev]{cleveref}
\usepackage{graphicx}
\usepackage{subcaption}

\DeclareMathOperator*{\argmax}{argmax}

\title{
Breeding Programs Optimization with Reinforcement Learning
}

\author{%
  Omar G. Younis \\
  ETH Zurich\\
  \texttt{omar.g.younis@gmail.com} \\
  \And
  Luca Corinzia\\
  ETH Zurich\\
  \texttt{luca.corinzia@inf.ethz.ch} \\
  \And
  Ioannis N. Athanasiadis\\
  Wageningen University and Research \\
  \texttt{ioannis.athanasiadis@wur.nl} \\
  \And
  Andreas Krause\\
  ETH Zurich\\
  \texttt{krausea@ethz.ch} \\
  \And
  Joachim M. Buhmann\\
  ETH Zurich\\
  \texttt{jbuhmann@inf.ethz.ch} \\
  \And
  Matteo Turchetta\\
  ETH Zurich\\
  \texttt{matteo.turchetta@inf.ethz.ch} \\
}

\begin{document}

\maketitle

\begin{abstract}
Crop breeding is crucial in improving agricultural productivity while potentially decreasing land usage, greenhouse gas emissions, and water consumption. However, breeding programs are challenging due to long turnover times, high-dimensional decision spaces, long-term objectives, and the need to adapt to rapid climate change.
This paper introduces the use of Reinforcement Learning (RL) to optimize simulated crop breeding programs. 
RL agents are trained to make optimal crop selection and cross-breeding decisions based on genetic information. 
To benchmark RL-based breeding algorithms, we introduce a suite of Gym environments.
The study demonstrates the superiority of RL techniques over standard 
practices in terms of genetic gain when simulated in silico using real-world genomic maize data.
\end{abstract}

\section{Introduction}
\label{introduction}
\looseness=-1
Crop breeding programs aim to create new cultivars with desired traits by controlled mating among individuals in a population \cite{breeding-schemes}. Crop breeding played a vital role in the \textit{Green Revolution} that took place in the 20th century, which resulted in a threefold increase in plant-based food production within 50 years, while only raising land usage by 30\% \cite{green-revolution}. However, current growth trends do not meet the forecasted 2050 demand. Moreover, evidence shows that yields for staple crops are plateauing or degrading mainly due to climate change and soil degradation \cite{growth-trends}. 

The decrease in genotyping cost has produced abundant crop genomic data, unveiling novel opportunities to adapt crop genetics for these adverse conditions. Recent advancements have focused on predicting phenotype traits from genomic data for quicker and cheaper crop selection without expensive and slow trials \cite{Meuwissen2001-ut, crossa2017genomic, CIMMYT}. However, relying solely on estimated traits can lead to low-diversity pools and compromise long-term breeding program success. 

Data-driven methods for decision-making can fill this gap and allow breeding programs to retain genetic diversity and adapt to long-term objectives, e.g., climate adaptation. To use these methods to optimize crop breeding, we can frame a breeding program as a sequential decision-making problem where, at each breeding generation, the breeder takes some actions, e.g., which and how many plants to select, and how to cross them. Crossing two plants produces offspring with a mixture of genes from the two parents. The process is repeated for several generations to deploy a cultivar with desired traits. From this perspective, we can optimize design choices of breeding programs with Reinforcement Learning (RL) \cite{sutton2018reinforcement}, a paradigm to address sequential decision-making problems.

\textbf{Contributions:} In this paper, we develop a framework to optimize the design choices of breeding programs, with three main contributions.
(1) In \cref{sec:problem_definition}, we model a breeding program as a Markov Decision Process (MDP), allowing us to leverage techniques from the RL literature to adaptively optimize the design choices. 
(2) For this scope, we present in \cref{sec:environments} a package containing multiple RL environments that simulate breeding programs with different sets of actions and observations. 
(3) In \cref{sec:experiments}, we show the effectiveness of the approach in selecting plants based on a learned score; this approach outperforms the standard genomic selection based on the estimated traits. Please note that the results presented are based on simulations and may differ in real-world applications.

\textbf{Related work:} 
Machine learning models have been applied in breeding programs in various ways \cite{genes14040777}, however, often for trait prediction in genomic assisted breeding \cite{cattle-breeding, svm_gp}. 
To the best of our knowledge, RL has been only used in the context of breeding in \cite{rl-budget-allocation}, where the authors optimize the population size at every step (generation) of the breeding program, while other decisions (e.g., genomic prediction function) are considered fixed.
\section{Problem definition}
\label{sec:problem_definition}
In reinforcement learning, sequential decision-making problems are formalized as Markov decision processes (MDPs) \cite{puterman2014markov}, i.e. a tuple $(S, \rho, \mathcal{A}, \mathcal{P}, \mathcal{R}, \gamma)$, where: $S$ is the set of all possible states; $\rho$ is the probability distribution of the initial state $s_0$; $\mathcal{A}: S \to A$ maps each state $s$ to the possible actions in $s$; $\mathcal{P}$ denotes the transition probability $\mathbb{P}(s_{t+1}\mid s_t,a_t)$ with $s_t,s_{t+1}\in S,a_t\in\mathcal{A}(s_t)$; $\mathcal{R}: S \times A \to \mathbb{R}$ is the reward function that defines the desirability of the state-action pair; $\gamma$ (between 0 and 1) balances the preference for immediate versus long-term rewards.\\
A policy $\pi$ is a function that maps a state $s_t$ to a distribution over valid actions in the set $\mathcal{A}(s_t)$.
RL algorithms aim to find a policy that maximizes the expected return, i.e.,
\begin{equation}
    \pi^* = \argmax_{\pi \in \Pi} \mathbb{E}_{\mathcal{P}, \pi, \rho} \left[\sum_{t=0}^T \gamma^t \mathcal{R}(s_t, a_t)\right]
\end{equation}
where $T$ is the length of the episode, and $\Pi$ is the set of possible policies.
In the case of breeding programs, at each generation $t$, the breeder performs arbitrary crosses obtaining a new population.
We define the state at time step $t$ as the set of genomes of the population
\begin{equation}
s_t = \{g^{(t)}_1, g^{(t)}_2, ..., g^{(t)}_{n_t}\} \in S
\end{equation} where $g^t_i$ represent the genome of the $i$-th individual at generation $t$, and $n_t$ is the population size at generation $t$. We explain in the next section how we represent a genome $g^t_i$ in silico.
Let's denote the set of indices of the population at step $t$ as $[n_t] = \{1, ..., n_t\}$.
An action at timestep $t$ consists of making an arbitrary number of crosses from pairs of plants from the set $[n_t]$, i.e. 
\begin{equation}
\label{eq:actions}
\mathcal{A}(s_t) = \{(e_1, ..., e_l) \mid l\geq 1, e_i \in [n_t] \times [n_t]\} \ .
\end{equation}
Given such action, the population at the subsequent timestep has size $l$, hence $n_{t+1} = l$. The transition probability $\mathbb{P}(s_{t+1} | s_{t}, a_{t})$ describes the biological process of the cross (genetic recombination) according to action $a_t$. 
Since a single breeding cycle can take several years and RL requires many interactions with the environment, we simulate the transition probability \textit{in silico}.

\section{Genetic simulation}
\label{sec:genetic_simulation}
Our experiments focus on crops with two full sets of chromosomes, called diploid crops (the setting is easily extensible to other ploidy).
Among all the nucleotides in a chromosome, we are interested in those varying among the same type of crop, called Single Nucleotide Polymorphisms (SNPs).
In particular, we are interested in the SNPs affecting the trait of interest. We assume every SNP has two possible configurations, hence they can be represented by a boolean array of size $(m, 2)$, where $m$ is the total number of SNPs and the second axis is due to diploidy. When two plants are mated, the offspring inherits one chromosome set from each parent. The inherited chromosome is formed by a stochastic biological process called genetic recombination, which involves the mixing of genetic material from the parental chromosome.
Simulating this process involves processing large arrays and can be slow. We rely on \textsc{ChromaX} \cite{younis2023chromax}, a performance-oriented breeding simulator based on JAX \cite{jax2018github}, which allows us to easily parallelize the operations on GPU(s).

The estimated trait value (e.g. yield) is computed using a linear regression model trained on real data:
\begin{equation}
    \hat{y} = \sum_{i=1}^m w_i \cdot (g_{i, 1} + g_{i, 2})
\end{equation}
where $m$ is the total number of SNPs, $w_i$ is the linear regressor weight for the SNPs $i$, and $g_{i, 1}, g_{i, 2}$ are the boolean values of the $i$-th SNPs for the first and second chromosome, respectively.

\section{Gym environments}
\label{sec:environments}
We introduce a set of Gym environments \cite{openai-gym} to train RL agents to learn breeding program design choices at different levels of complexity.
Utilizing the vectorization capabilities of JAX \cite{jax2018github}, the environments parallelize and potentially distribute computations across hardware accelerators.

\textsc{\textbf{BreedingGym:}}
The base environment, where the agent observes the genomes of the population and can perform all possible crosses. The observation space is a variable-sized sequence of arrays. The action size is also dynamic, represented as a sequence of integer pairs as defined in \cref{eq:actions}. During initialization, the user can specify the genetic data, the horizon $T$, and whether the agent should be rewarded during the episode or solely at the conclusion. 
Rewards are computed using aggregation functions (e.g., max, mean) of trait values over the population, estimated from the genetic data.

\textsc{\textbf{SimplifiedBreedingGym:}}
This environment fixes the size of the population, resulting in static dimensions of observations and actions. The observation is a dictionary containing, for each plant, the yield and the genome correlation coefficient. The action is a dictionary containing the number of plants to select based on a given trait and how many random crosses to perform on them. To keep the number of plants constant at each generation, the crosses can produce more than one offspring.

\textsc{\textbf{SelectionScores:}}
Genomic selection in breeding often focuses on trait values when selecting plants (e.g., higher yields). Yet, this might ignore the potential of certain plants with poor current performance. Selection scores like optimal haploid value \cite{ohv} are based on this intuition. 
This environment is aimed at learning a non-myopic selection score. The observation is the genome, while the action is an array containing a score for each plant. The $k$ top-scoring plants are chosen and crossed randomly. The user fixes the population size, the value $k$, and the number of crosses $l$.

\textsc{\textbf{PairScore:}}
In this environment, the agent assigns a score for every possible cross. This allows for more degrees of freedom in choosing which parents have more potential if mated together. Thus, for a population of size $n$, the action will be a matrix of size $n \times n$. Then, $n$ crosses are selected according to the score matrix, where $n$ is fixed by the user.
\section{Experiments}
\label{sec:experiments}
\looseness=-1
We present here the results obtained on the \textsc{SelectionValues} environment with fixed population size $n_t = 200$. At every step, we select $k=20$ plants based on the scores in the action and we make 10 random crosses with 20 offspring each. Additionally, we fix the number of generations to $T=10$, and we reward the agent based only on the estimated trait of the best individual of the last generation. 

\textbf{Data and observations:} We use maize genetic data from \cite{look-ahead} which contains the SNPs values of a population along with the linear regressor weights $w$, to predict the Shoot Apical Meristem (SAM) volume \cite{Ha2010-pu}, which we use as a proxy for yield. In every episode, we sample 200 plants to compose the initial population. To reduce memory usage and speed up the simulation, we randomly subset a total of $m=1000$ SNPs. Consequently, the observation is an array with shape $(200, 1000, 2)$.
To ease training, we process the observation by multiplying it with $w$. In this way, the estimated SAM volumes can be obtained by summing the values over the last two axes of the observation array.

\textbf{Algorithm:} We use the Proximal Policy Optimization \cite{schulman2017proximal} algorithm implemented in StableBaselines3 \cite{stable-baselines3}.
To make the policy invariant to the permutation of plants in the population, we use a neural network that processes the plants independently and outputs the score, i.e. a scalar value.

\looseness=-1
\textbf{Architecture:} To keep SNPs adjacency information while reducing the number of policy parameters, we process the observation with a 1-dimensional convolutional neural network. We also include information about the generation number as the percentage of completion of the episode. More information about the architecture can be found in \cref{appendix:architecture}.

\begin{figure}
\begin{subfigure}{.4\textwidth}
\includegraphics[trim=2mm 3mm 0 3mm, clip, width=.95\columnwidth]{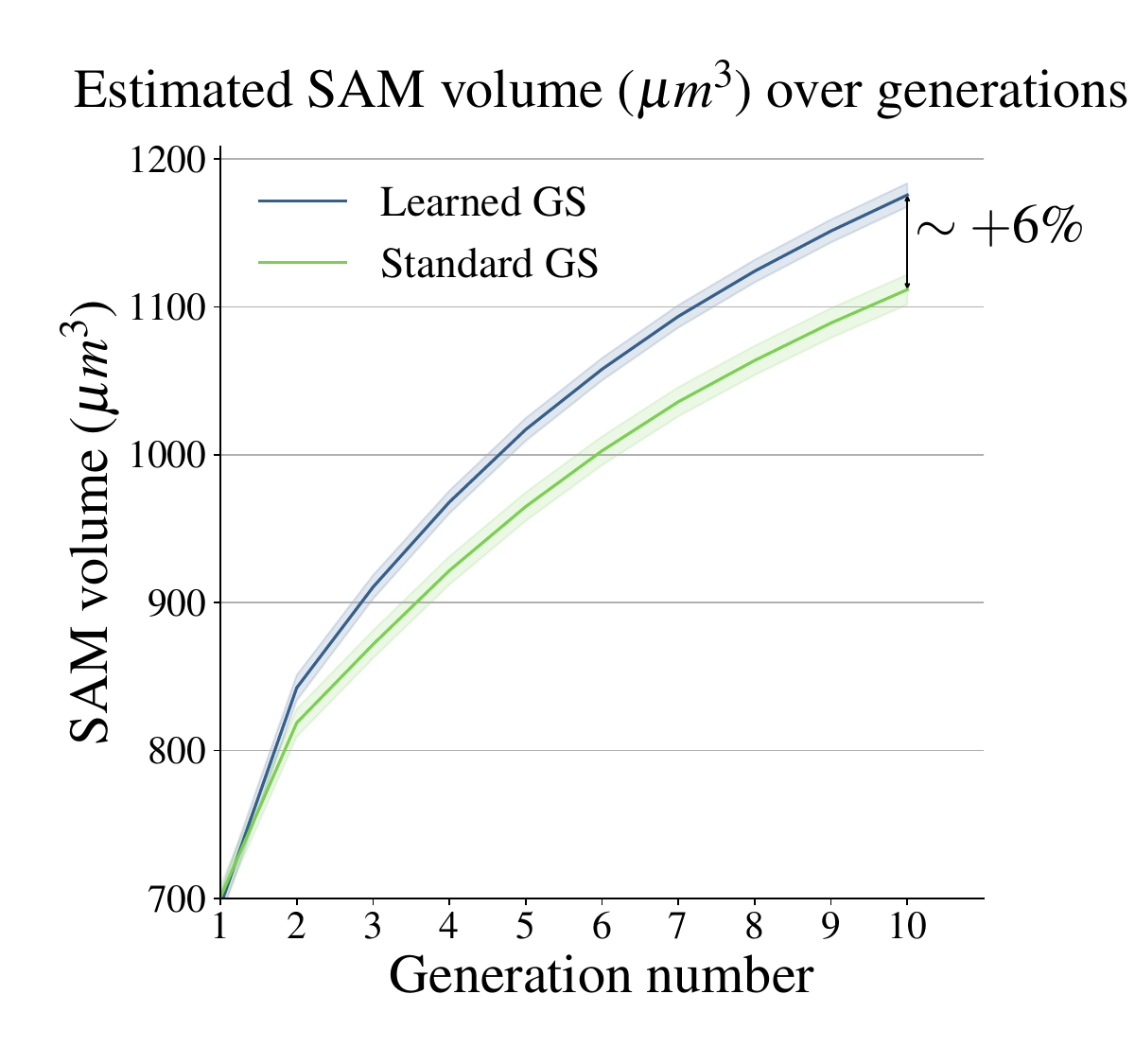}
\end{subfigure}
\begin{subfigure}{.6\textwidth}
\includegraphics[trim=0mm 13mm 20mm 0mm, clip, width=\columnwidth]{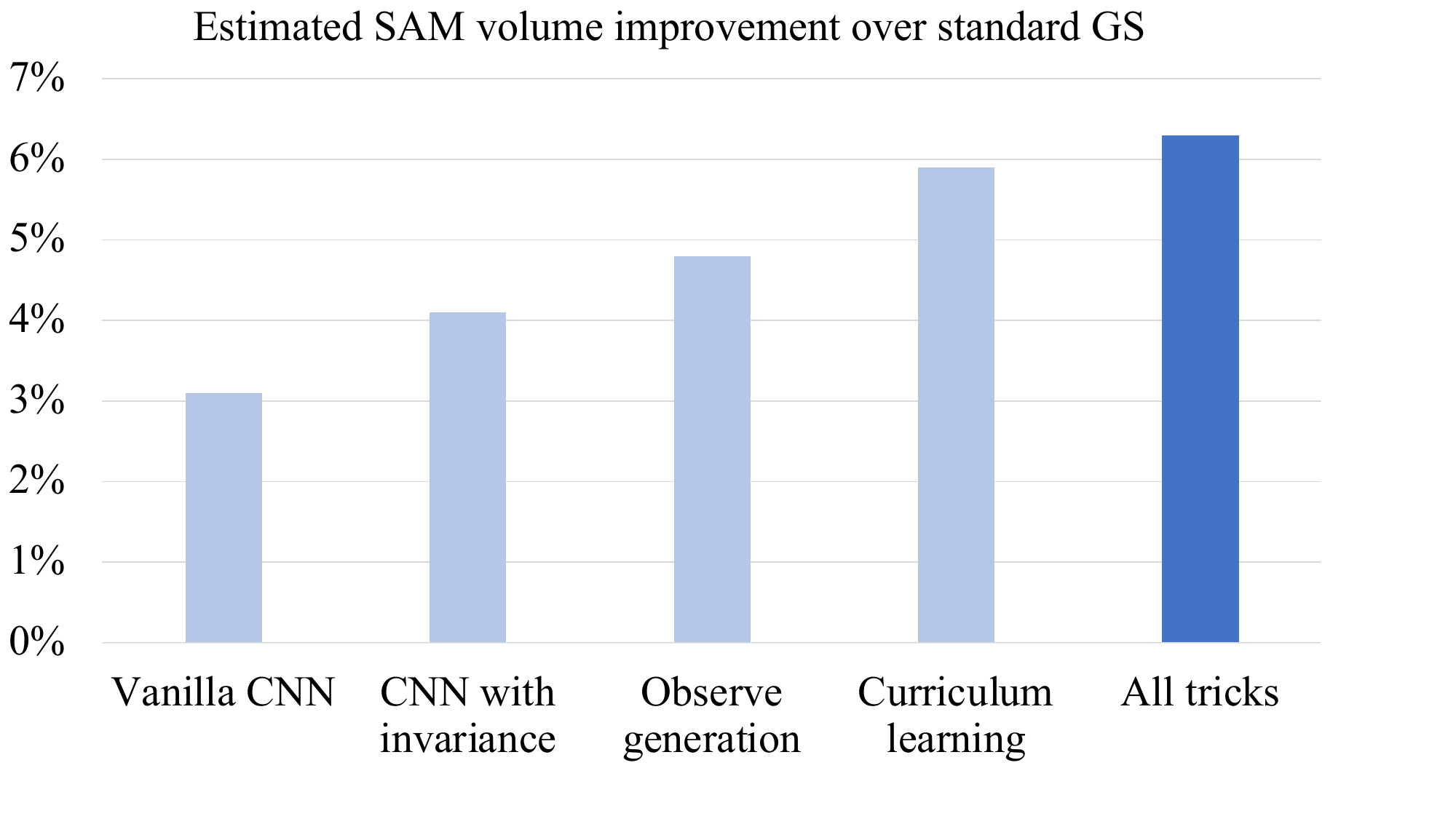}
\end{subfigure}
\caption{
\textit{(On the left)} Estimated SAM volume in $\mu m^3$ during the breeding program with standard and learned GS, averaged across 100 trials, with shaded area indicating standard error.\\
\textit{(On the right)} Estimated SAM volume percentage increase by applying the tricks explained in this section. Note: on "Curriculum learning" we also let the agent observe the generation number.}
\label{fig:sam_over_generations}
\end{figure}

\textbf{Training and results:} To help the agent handle sparse reward, we do curriculum learning \cite{narvekar2020curriculum} by gradually extending the time horizon from $T=3$ to $T=10$ during training.
We trained for 7 million generations on a \textit{NVIDIA GeForce RTX 2080 Ti}, taking around 24 hours. 
\cref{fig:sam_over_generations} shows the increase in the estimated SAM volume during the 10 generations of the breeding program and the effectiveness of the employed tricks. 
Our genomic selection method (Learned GS) obtains an estimated 1176$\mu m^3$ SAM volume, around 6\% higher than the one obtained by standard genomic selection (Standard GS).

\section{Discussion}
\label{sec:discussion}
\looseness=-1
This paper introduces a new approach to optimize breeding programs. We formulate the problem as a Markov decision process, making techniques from the RL literature applicable to this challenge. We show the effectiveness of the method by training an RL agent to select plants during the breeding program. 
To foster research in this field, we publish a collection of Gym environments  that allow researchers to optimize various aspects of breeding programs. The genetic processes are simulated, enabling swift and cheap exploration of potential solutions. However, it is important to acknowledge the gap between simulated processes and reality, which can lead to considerable performance degradation.

Beyond the societal relevance of breeding scheme optimization, these environments present a set of interesting challenges for RL research. Specifically, the observation space can be very large as well as the action space, which scales combinatorially with the population size. Moreover, their size is in general dynamic and can be decided by the agent; in this case, we typically need to introduce constraints that reflect real-world costs and budgets. Moreover, the MDP is factored \cite{fmdp} as the state is composed of weakly connected components 
(the genomes of the plants). Finally, the reward can be non-stationary as the regressor model is continuously trained with new data.

We hope this work can foster collaborations across artificial intelligence and the agriculture community, and drive further advancements in the optimization of breeding programs. Our efforts seek to make a tangible impact on agricultural practices and contribute to addressing global challenges related to food security and sustainability.

\newpage
\bibliography{references}

\begin{thebibliography}{10}

\bibitem{breeding-schemes}
Covarrubias-Pazaran, G., Z.~Gebeyehu, D.~Gemenet, et~al.
\newblock Breeding schemes: What are they, how to formalize them, and how to
  improve them?
\newblock \emph{Frontiers in Plant Science}, 12, 2022.

\bibitem{green-revolution}
Pingali, P.~L.
\newblock Green revolution: Impacts, limits, and the path ahead.
\newblock \emph{Proceedings of the National Academy of Sciences},
  109(31):12302--12308, 2012.

\bibitem{growth-trends}
Ray, D.~K., N.~Ramankutty, N.~D. Mueller, et~al.
\newblock Recent patterns of crop yield growth and stagnation.
\newblock \emph{Nat Commun}, 3:1293, 2012.

\bibitem{Meuwissen2001-ut}
Meuwissen, T.~H., B.~J. Hayes, M.~E. Goddard.
\newblock Prediction of total genetic value using genome-wide dense marker
  maps.
\newblock \emph{Genetics}, 157(4):1819--1829, 2001.

\bibitem{crossa2017genomic}
Crossa, J., P.~P{\'e}rez-Rodr{\'\i}guez, J.~Cuevas, et~al.
\newblock Genomic selection in plant breeding: methods, models, and
  perspectives.
\newblock \emph{Trends in plant science}, 22(11):961--975, 2017.

\bibitem{CIMMYT}
Dreisigacker, S., J.~Crossa, P.~Pérez-Rodríguez, et~al.
\newblock Implementation of genomic selection in the cimmyt global wheat
  program, findings from the past 10 years.
\newblock \emph{Crop Breeding, Genetics and Genomics}, 3(2):e210004, 2021.
\newblock Doi: 10.20900/cbgg20210005.

\bibitem{sutton2018reinforcement}
Sutton, R.~S., A.~G. Barto.
\newblock \emph{Reinforcement learning: An introduction}.
\newblock MIT press, 2018.

\bibitem{genes14040777}
Yoosefzadeh~Najafabadi, M., M.~Hesami, M.~Eskandari.
\newblock Machine learning-assisted approaches in modernized plant breeding
  programs.
\newblock \emph{Genes}, 14(4), 2023.

\bibitem{cattle-breeding}
Yu, T., W.~Zhang, J.~Han, et~al.
\newblock An ensemble learning approach for predicting phenotypes from
  genotypes.
\newblock In \emph{2021 20th International Conference on Ubiquitous Computing
  and Communications (IUCC/CIT/DSCI/SmartCNS)}, pages 382--389. 2021.

\bibitem{svm_gp}
Zhao, W., X.~Lai, D.~Liu, et~al.
\newblock Applications of support vector machine in genomic prediction in pig
  and maize populations.
\newblock \emph{Frontiers in Genetics}, 11, 2020.

\bibitem{rl-budget-allocation}
Moeinizade, S., G.~Hu, L.~Wang.
\newblock A reinforcement learning approach to resource allocation in genomic
  selection.
\newblock \emph{Intelligent Systems with Applications}, 14:200076, 2022.

\bibitem{puterman2014markov}
Puterman, M.~L.
\newblock \emph{Markov decision processes: discrete stochastic dynamic
  programming}.
\newblock John Wiley \& Sons, 2014.

\bibitem{younis2023chromax}
Younis, O.~G., M.~Turchetta, D.~Ariza~Suarez, et~al.
\newblock Chroma{X}: a fast and scalable breeding program simulator, 2023.

\bibitem{jax2018github}
Bradbury, J., R.~Frostig, P.~Hawkins, et~al.
\newblock {JAX}: composable transformations of {P}ython+{N}um{P}y programs,
  2018.

\bibitem{openai-gym}
Brockman, G., V.~Cheung, L.~Pettersson, et~al.
\newblock Openai gym, 2016.

\bibitem{ohv}
Daetwyler, H.~D., M.~J. Hayden, G.~C. Spangenberg, et~al.
\newblock Selection on optimal haploid value increases genetic gain and
  preserves more genetic diversity relative to genomic selection.
\newblock \emph{Genetics}, 200(4):1341--1348, 2015.

\bibitem{look-ahead}
Moeinizade, S., G.~Hu, L.~Wang, et~al.
\newblock Optimizing selection and mating in genomic selection with a
  {Look-Ahead} approach: An operations research framework.
\newblock \emph{G3 (Bethesda)}, 9(7):2123--2133, 2019.

\bibitem{Ha2010-pu}
Ha, C.~M., J.~H. Jun, J.~C. Fletcher.
\newblock Shoot apical meristem form and function.
\newblock \emph{Curr Top Dev Biol}, 91:103--140, 2010.

\bibitem{schulman2017proximal}
Schulman, J., F.~Wolski, P.~Dhariwal, et~al.
\newblock Proximal policy optimization algorithms, 2017.

\bibitem{stable-baselines3}
Raffin, A., A.~Hill, A.~Gleave, et~al.
\newblock Stable-baselines3: Reliable reinforcement learning implementations.
\newblock \emph{Journal of Machine Learning Research}, 22(268):1--8, 2021.

\bibitem{narvekar2020curriculum}
Narvekar, S., B.~Peng, M.~Leonetti, et~al.
\newblock Curriculum learning for reinforcement learning domains: A framework
  and survey.
\newblock \emph{The Journal of Machine Learning Research}, 21(1):7382--7431,
  2020.

\bibitem{fmdp}
Boutilier, C., R.~Dearden, M.~Goldszmidt.
\newblock Exploiting structure in policy construction.
\newblock In \emph{Proceedings of the 14th International Joint Conference on
  Artificial Intelligence - Volume 2}, IJCAI'95, page 1104–1111. Morgan
  Kaufmann Publishers Inc., San Francisco, CA, USA, 1995.

\end{thebibliography}
\bibliographystyle{references}

\newpage
\appendix
\section{Architecture details}
\label{appendix:architecture}

\begin{figure}[h]
\includegraphics[trim=2mm 3mm 0 3mm, clip, width=.95\columnwidth]{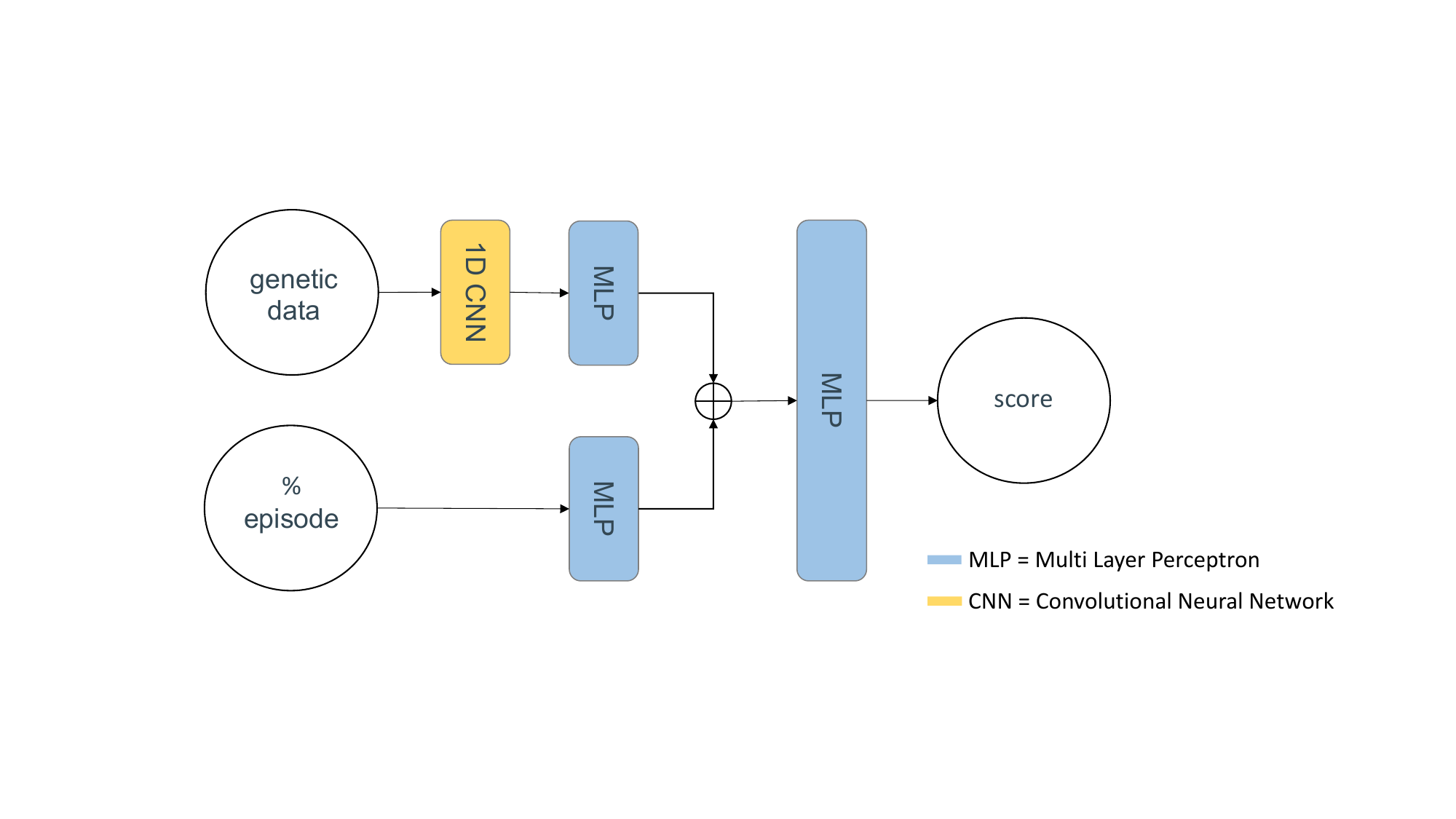}
\caption{Representation of the policy net. The final score is a scalar value that is used by the environment to select the plants to cross.}
\label{fig:policy_net}
\end{figure}

In the following, we explain the policy architecture that we used in the experiments presented in \cref{sec:experiments}.
We represent the policy network in \cref{fig:policy_net}.

The network processes each plant in an episode independently and it comprises two components: a feature extractor for processing the genetic array data, yielding 64 features, and another network responsible for generating the scalar value, which relies on these features and the generation number. As we use Proximal Policy Optimization, we additionally utilize a value network that shares the feature extractor with the action network.

The first convolutional layer has 64 kernels of length 32 and stride 8, and the second one with 16 kernels of length 8 and stride 2. With the input having shape $(1000, 2)$ and these parameters, the flattened output has a size of 928. 
This output is then processed by a Multi-Layer Perceptron layer that maps the 928 array to 64 features.
As there is no inherent semantic order between the two chromosomes we ensure the network's invariance to the permutation of the two channels. To achieve this, we average the output of the feature extractor considering both possible orders. 

Additionally, we incorporate information about the generation number to generalize better when changing the time-horizon. We do so by creating a mapping from the completion percentage of the breeding program to 16 features. These 16 features are concatenated with the previously obtained 64 features. We use two different mapping for the action and the value networks.

The final step involves processing these combined 80 features with two MLPs (one for the action and one for the value) with 32 hidden layers and scalar outputs. Finally, to determine the value of the current state, we average the outputs (one for each plant) produced by the value head.

\end{document}